# Development of an automated Red Light Violation Detection System (RLVDS) for Indian vehicles

Satadal Saha [1], Subhadip Basu [2][*], Mita Nasipuri [2], Dipak Kumar Basu [#][2]
[#] *AICTE Emeritus Fellow*
[1] *CSE Department, MCKV Institute of Engineering, Howrah, India*
[2] *CSE Department, Jadavpur University, Kolkata, India*
[*] *Corresponding author. Email: subhadip@ieee.org*

**Abstract**

*Integrated Traffic Management Systems (ITMS) are now implemented in different cities in India to primarily address the concerns of road-safety and security. An automated Red Light Violation Detection System (RLVDS) is an integral part of the ITMS. In our present work we have designed and developed a complete system for generating the list of all stop-line violating vehicle images automatically from video snapshots of road-side surveillance cameras. The system first generates adaptive background images for each camera view, subtracts captured images from the corresponding background images and analyses potential occlusions over the stop-line in a traffic signal. Considering round-the-clock operations in a real-life test environment, the developed system could successfully track 92% images of vehicles with violations on the stop-line in a "Red" traffic signal.*

## 1. Introduction

Integrated automated traffic management systems are now being implemented across different major cities in India. Primary objective of such systems is to track down vehicles that violated traffic regulations using surveillance cameras and intelligent image analytic softwares. The current work on development of an automated *Red Light Violation Detection System (RLVDS)* may be considered as key module of the overall *Integrated Traffic Management System (ITMS)*.

The objective of the current research work is to develop an automated system to localize vehicles violating the traffic "Red" light signal and capture the images of such vehicles with date, time and location information.

Now-a-days in most of the road crossings the traffic is controlled by automatic signaling system. In general red, yellow and green colored lights are used for interpretation of three types of signals for traffic controlling operations. Green light signals to start a stopped vehicle, yellow light signals a moving vehicle to slow down the speed and red light signals the vehicle to stop. In any road crossing when red light is shown to a lane, the signal conveys the message to the vehicles rushing towards the crossing to stop immediately. To make the procedure more systematic and more convenient a uniform thick white line is drawn in each lane before the crossing. This line is commonly known as *stop-line*. Each vehicle coming towards the crossing must stop before this line if red signal is seen by it. Even if the front wheel of the vehicle touches the stop-line partially then also it is decided as a red light violating vehicle. Stop-line is usually placed perpendicular to the direction of flow of traffic and is placed in the plane of the road.

The task for detecting the stop-line violating vehicles (on a "Red" traffic signal) is done by the traffic police manually. In a busy schedule with thousands of vehicles passing through the crossing it may not always be possible to manually generate a full list of all the violating vehicles. This has been the primary motivation behind the work presented in this paper. We have installed multiple cameras by the road-side at a certain height to capture the image automatically at a regular interval of time, processed all the images captured by the camera and generated a list of all violating vehicles. A well designed interface has been developed to get full list of images automatically updated at regular interval of time.

Not much work has been done on automatic localization of "Red" light violating vehicles specifically. Most of the existing automatic traffic monitoring systems [1-7] concentrate on localization of license plate regions and subsequent interpretation of the characters therein. The automatic License Plate Recognition (ALPR) techniques are predominantly designed for different developed countries with standardized vehicle license plates and strictly monitored integrated traffic management systems. However, in Indian scenario with minimal traffic monitoring these systems fail miserably. In general, most of the works on ALPR systems [1-4] apply edge based features for localizing standardized license plate regions. Some of these works [2, 5, 6] captures the image of a vehicle carefully placed in front of a camera occupying the full view of it and taking a clear image of the license plate. Unfortunately in the practical scenario there may be multiple vehicles of different types in a single scene along with partial occlusions of the vehicles and also the license plates from other objects, where the above methods do not work. In one of the earlier works [1], Rank filter is used for localization of license plate regions giving unsatisfactory result for skewed license plates. An analysis of Swedish license plate is done in [2] using vertical edge detection followed by binarisation. This does not give better result for non-uniformly illuminated plates. An exhaustive study of plate recognition is done in [3] for different European countries. In Greece the license plate uses shining plate. The bright white background is used as a characteristic for license plate in [4]. Spanish license plate is recognized in [5] using Sobel edge detection operator. It also uses the aspect ratio and distance of the plate from the center of the image as characteristics. But it is constrained for single line license plates. During the localization phase the position of the characters is used in [6]. It assumes that no significant edge lies near the license plate and characters are disjoint.

In course of the work, we have collected a dataset for the experiment, as discussed in section 2. Section 3 describes the procedure of the experiment and includes the subsections 3.1 to 3.3 explaining the methods for adaptive background image generation, background subtraction and identification of stop-line and stop-line violation detection respectively. In section 4 results of the experiment are shown and discussed. Section 5 gives the overall conclusion of the present work.

## 2. Collection of dataset

The dataset for the current experiment is collected as a part of a demonstration project on *Automated Red Light Violation Detection* system for a Government traffic monitoring authority of a major metro city in India. Three surveillance cameras were installed at an important road crossing in Kolkata at a height of around ten meters from the road surface. Two of these were static cameras with manually adjustable zoom and focus, fixed at a given orientation with the road surface. The third one was a auto-focus pan-tilt-zoom (PTZ) camera, which was used to pan the entire road width at six preset pan angles automatically at an interval of 3 seconds. All the surveillance cameras were synchronized with the traffic signaling system such that the camera captures the video snapshots only when the traffic signal is turned *RED*. All the cameras were focused on the *Stop-Line* to capture frontal images of vehicles violating the *Stop-Line* on a *RED* traffic signal.

The complete image dataset comprises of more than 15,000 surveillance video snapshots, captured over several days/nights in an unconstrained environment with varying outdoor lighting conditions, pollution levels, wind turbulences and vibrations of the camera. 24-bit color bitmaps were captured through CCD cameras with a frame rate of 25 fps and resolution of 704 x 576 pixels. Not all these video snapshots contain vehicle images with a clear view of license plate regions. For the current experiment, we have identified 500 images with complete license plate regions appearing in different orientations in the image frame.

## 3. Present work

Installation of the camera and the interfacing with it is a trivial task for successful and satisfactory completion of the rest of the work. As the camera can not be placed at the middle of a road at a very low level (as already mentioned it is placed at the side of a road at a convenient height), normally the stop-line makes an angle with the projection plane of the camera. Our present work is based on scanning the image along the stop-line and to detect whether there is any violation or not. For this purpose the value of the aforesaid angle is to be known in advance. As already mentioned that we have used 3 cameras (two fixed cameras and one PTZ camera) for collecting the images of vehicles coming through different lanes, the angles made by the stop-line are different for the three sets of images. In case of CAM1 and CAM2 the angle remains fixed but different for each of the firmly installed cameras, which helps us correcting the angle internally during the processing of the images captured with these cameras. The problem is more complicated in case of PTZ camera which spans the road for focusing different lanes of it and captures images for each lane. In this case the same stop-line makes different angles for different facing of the camera. So synchronization between the direction of camera and angle of correction is required in this case, i.e. for each discrete spanning angle (we have used six spanning angles to cover the whole width of the road) of the camera the angle made by the stop-line in the image should be known in advance.

The scheme for detection of red light violating vehicles is based on background subtraction technique. We may define the term background image as the image without any movable object in it. So an image consisting of road, pedestrian, wall, tree etc. may be called of as background image. If the background image is known in advance then the intrusion of any other object into the scene may be obtained by subtracting the non-background image from the background image. Fig. 1 depicts the flow of work.

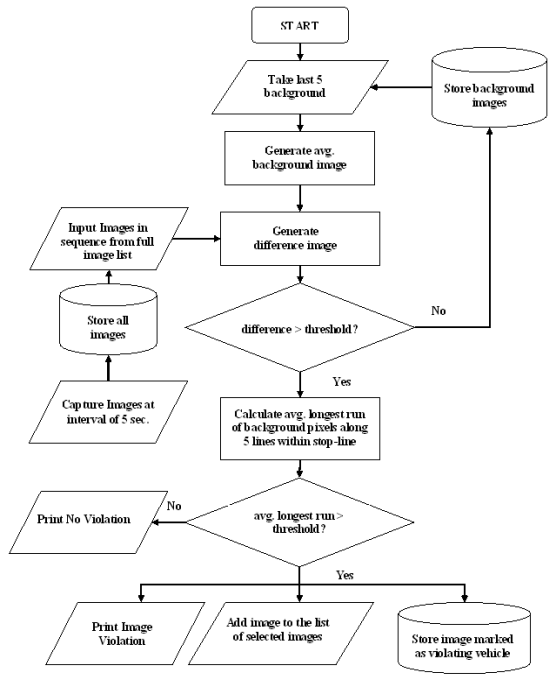

**Fig. 1. Flow chart of the scheme**

### 3.1. Adaptive background image generation

One key factor in the process is that the background changes automatically from morning to evening and from evening to night due to the variation of sun light and street light respectively. So to detect an object from an image captured at night, if the image is subtracted from the background image of morning then obviously wrong prediction of object will result. So there must be some mechanism implemented such that change in background may also be adapted with the change in time.

We have adopted an adaptive technique for gradual modification of background image. Starting from five initial background images, if for any successive image the difference with the background image is found to be very small (less than a predefined threshold value) then the image is considered as a background image and it is stored in the background image database. Once there is any modification in the background image database, the average background image is generated by averaging the last five background images. This average image is considered as a background image till next time the background is calculated again. In this way as soon as any new image is considered as a background image the average background image is recalculated and used for future work. Thus the averaging of the backgrounds has also become adaptive in nature. Fig. 2(a) and 2(b) shows the mean background images of the two static cameras. Fig. 2 (c-h) shows mean background images of the PTZ camera for six different pan angles.

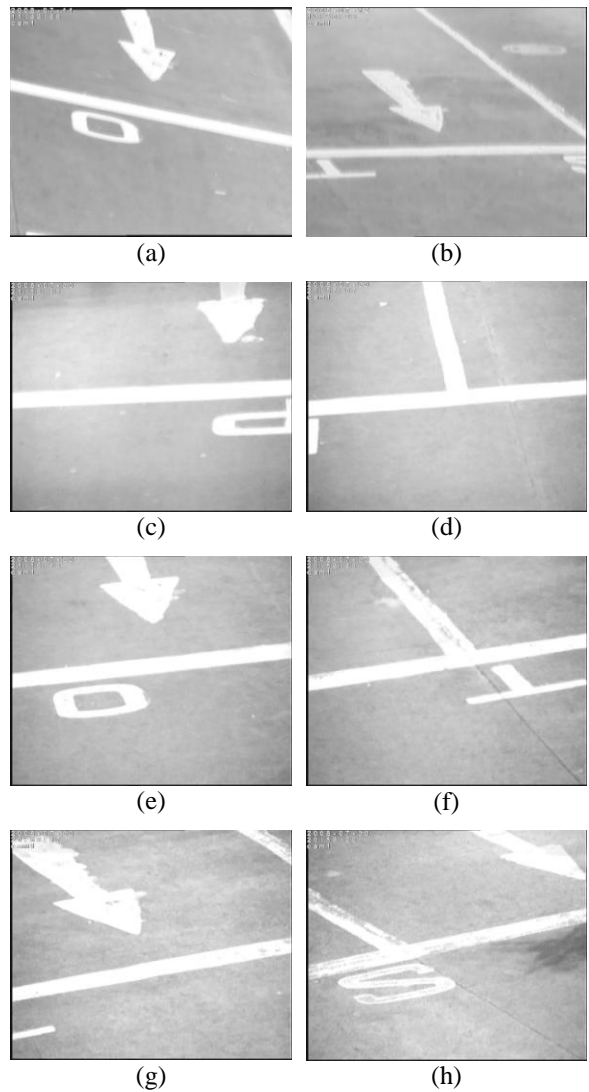

**Fig. 2(a-h). Mean background images of different surveillance cameras.**

### 3.2. Background subtraction and identification of stop-line

The three cameras send images at a regular interval of 3 seconds. These images are temporarily stored in the hard disk and a dynamic list is also prepared and stored in the disk. A process is written to automatically read the list of images written in a particular file and to read the images sequentially. A new image taken from the image list is always subtracted from the background image to find whether it is another background image or it is a non-background image. Fig. 3(a) shows a sample video snapshot captured by the first static camera and the corresponding subtracted image generated by subtracting it from the mean background image, as shown in fig. 2(a). In the difference image the gray values of all the pixels are added and then the summation is divided by the number of pixels to get the mean gray value over the image. If the difference image results a high mean gray value (greater than a predefined threshold $D_{th}$) then it is considered as a non-background image, i.e. there is a possibility of intrusion of an object in the image. Otherwise it is considered as a background image.

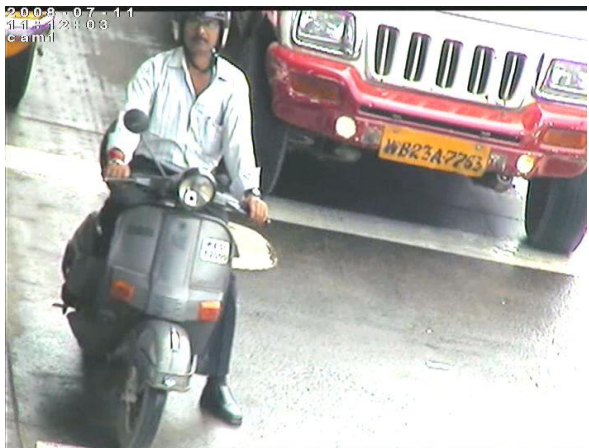

(a)

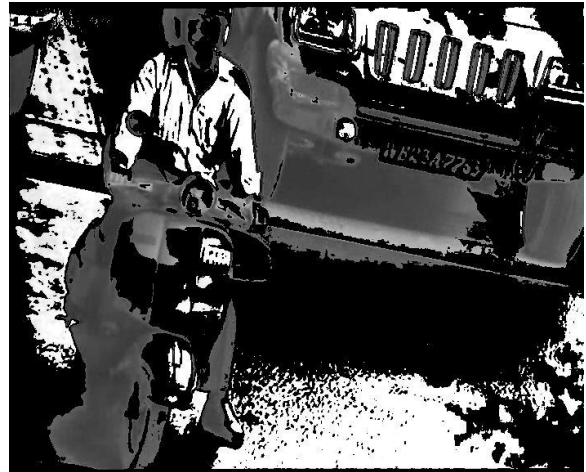

(b)

**Fig. 3(a). A sample vehicle image captured by one of the static cameras.**
**(b). Corresponding result of subtraction of the vehicle image from the mean background image shown in fig. 2(a).**

### 3.3. Stop-line violation detection

Any non-background image such as an image containing a vehicle when differenced with the background image produces black pixels throughout the image except where the movable objects are there in the former image. So in the difference image the white stop-line will also become black if it is not occluded by the vehicle. For detection of this occlusion we have hypothetically drawn five lines on the stop-line parallel to the longer edges of the stop-line, with a vertical gap of three pixels between them as shown in Fig. 4. In the difference image if all the pixels along the lines are black then it is decided that there is no occlusion and if any pixel is found to be non-black then it is decided that occlusion has occurred. Now, one problem regarding the generation of non-black pixels is noise in the difference image. This noise can generate randomly located non-black pixels in the image and some of them may lie on the hypothetical lines as well. But this should not be considered as the case of occlusion. So total number of continuous non-black pixels along the lines is important, not the presence of any isolated non-black pixel. The second problem is that though we expect continuous set of non-black pixels in case of occlusion of stop-line by a vehicle, this occlusion may also be occurred by the single or multiple passers by on the road crossing it laterally in front of the stop-line thereby occluding the stop-line. To solve this problem we have computed the longest run of non-background pixels along the hypothetical lines and calculated the mean of them. We here assume that a few persons can occlude the stop-line discretely but a vehicle (preferably four wheelers) will definitely occlude the stop-line continuously through a larger distance. Thus if the mean longest occlusion exceeds a predefined threshold $L_{th}$ then it is considered as a true occlusion made by a vehicle and the vehicle is considered as a stop-line violating or red light violating vehicle.

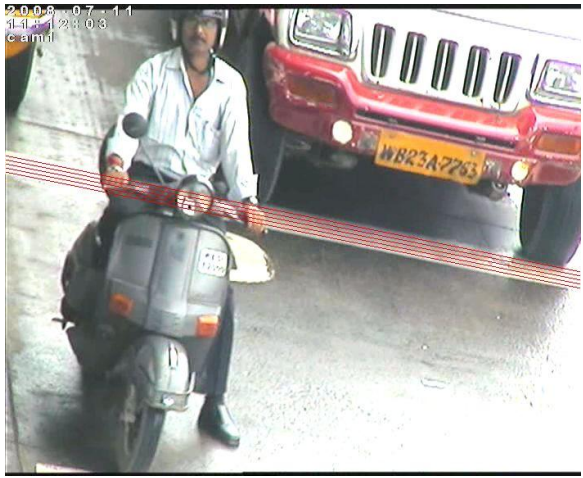

**Fig. 4. A sample violating vehicle image with potential occlusions on the stop-line (hypothetically marked with a band of red lines).**

## 4. Results and discussion

As discussed in section 2, more than 15000 video snapshots of resolution 704 x 576 are captured through multiple surveillance cameras. During the current experiment, 500 images were selected for the testing purpose of the technique presented in this paper. The experimental thresholds for computation of difference image ($D_{th}$) and estimation of true occlusion over the stop-line ($L_{th}$) were chosen as 70 and 140 respectively for the current experiment. We have developed a complete application software for RLVDS for identification of vehicle images violating the traffic signal. Fig. 5(a) shows the screenshots of the setup form of RLVDS to set the image path, select camera, estimate skew angle and tune different experimental thresholds. Fig. 5(b) shows the main application screen with potential images of the vehicles violating the stop-line. Fig. 5(c) shows sample screenshot of the violation-slip generated by the police department.

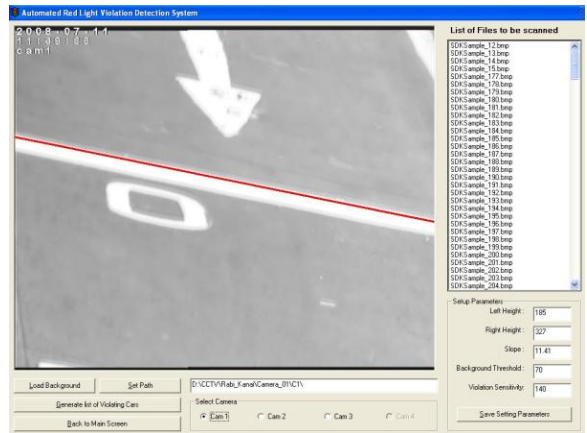

(a)

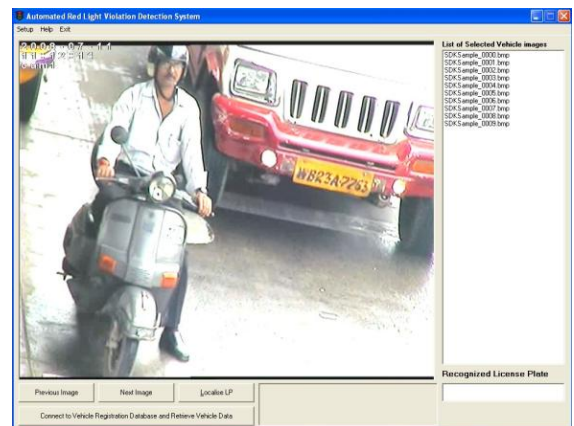

(b)

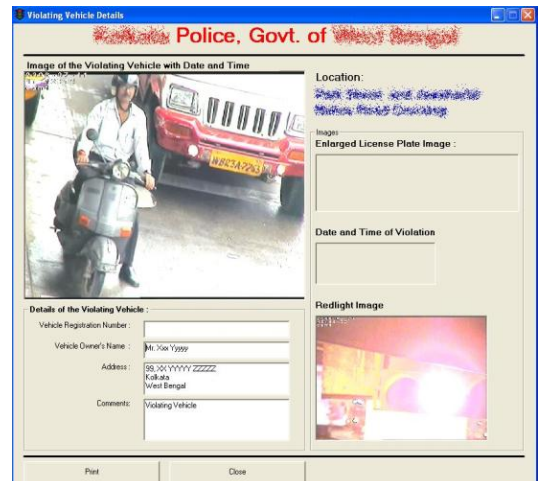

(c)

**Fig. 5. Snap shots of the application software.
(a) Setup screen with a list of all vehicle images
(b) Main application screen with a list of potentially violating vehicle images
(c) A sample violation-slip generated by the police department**

For computation of accuracy of the current technique, we have considered two different categories of error situations. Firstly, inaccurate tracking of non-violating vehicle images (in a traffic signal) and secondly, non-detection of actual violating vehicles. The overall observed error rate in this experiment is 8%. For all such vehicle images, that could be tracked perfectly during any possible *traffic-signal* violations, were identified as *true positive* cases. As observed from the experimentation on the collected dataset of 500 image samples of vehicle images the *true positive* accuracy is 92%. Major reason behind erroneous tracking of vehicle images is occlusion on the stop-line by pedestrians, speeding vehicles over the stop-line, erratic changes in outdoor lighting conditions and possible vibrations of the surveillance cameras.

## 5. Conclusion

In our present work we have developed a complete application for generating the list of all stop-line violating vehicles automatically from a full list of captured images. The technique successfully identifies vehicle images in 92% cases. All the images we have used here as data are all real life images and the high rate of success of our work directs that this algorithm can easily be implemented commercially in large scale at different crossings in the city to generate a full and centralized list of stop-line violating vehicles throughout the day and night with a minimum manual controlling operation. The current work can also be extended to localize the potential license plate regions from the list of violating vehicles and subsequently interpret the characters appearing on the license plate using an *Optical Character Recognition* module.

## 6. Acknowledgement